\documentclass[runningheads]{llncs}
\usepackage[T1]{fontenc}
\usepackage{graphicx}
\usepackage{hyperref}
\usepackage{color}

\usepackage{tabularx}
\usepackage{caption}
\usepackage{booktabs}
\usepackage{svg}
\usepackage{amsmath}

\begin{document}

\def\methodName{VISAGE}
\title{\methodName{}: Video Synthesis using Action Graphs for Surgery}
\titlerunning{VISAGE}
%
%
\author{Yousef Yeganeh\inst{1,2} \and Rachmadio Lazuardi\inst{1} \and Amir Shamseddin\inst{1} \and Emine Dari\inst{1} \and Yash Thirani\inst{1} \and Nassir Navab\inst{1,2,3} \and Azade Farshad\inst{1,2}}
\authorrunning{Yousef Yeganeh et al.}
\institute{Technical University of Munich \and Munich Center for Machine Learning \and Johns Hopkins University}
\maketitle              
\begin{abstract}
Surgical data science (SDS) is a field that analyzes patient data before, during, and after surgery to improve surgical outcomes and skills. However, surgical data is scarce, heterogeneous, and complex, which limits the applicability of existing machine learning methods. In this work, we introduce the novel task of future video generation in laparoscopic surgery. This task can augment and enrich the existing surgical data and enable various applications, such as simulation, analysis, and robot-aided surgery. Ultimately, it involves not only understanding the current state of the operation but also accurately predicting the dynamic and often unpredictable nature of surgical procedures. Our proposed method, \textbf{\methodName{}} (\textbf{VI}deo \textbf{S}ynthesis using \textbf{A}ction \textbf{G}raphs for Surg\textbf{e}ry), leverages the power of action scene graphs to capture the sequential nature of laparoscopic procedures and utilizes diffusion models to synthesize temporally coherent video sequences. \methodName{} predicts the future frames given only a single initial frame, and the action graph triplets. By incorporating domain-specific knowledge through the action graph, VISAGE ensures the generated videos adhere to the expected visual and motion patterns observed in real laparoscopic procedures. The results of our experiments demonstrate high-fidelity video generation for laparoscopy procedures, which enables various applications in SDS.
\end{abstract}

\section{Introduction}
\label{sec:intro}
Surgical data science (SDS) is an emerging field that focuses on the quantitative analysis of pre-, intra-, and postoperative patient data \cite{maier2017surgical,ozsoy20224d}; it can help to decompose complex procedures, train surgical novices, assess outcomes of actions, and create predictive models of surgical outcomes \cite{maier2022surgical,murali2023latent,holm2023dynamic,köksal2024sangriasurgicalvideoscene}. In recent years, there have been multiple works on surgical action recognition \cite{nwoye2020recognition,nwoye2021rendezvous,sharma2023rendezvous}. However, SDS faces several challenges, such as the scarcity and heterogeneity of surgical data, the variability and complexity of surgical processes, and the ethical and legal issues of data collection and sharing. To overcome these challenges, we introduce the novel task of surgical video generation, which aims to synthesize realistic and diverse videos of surgical procedures that can be used for various applications, such as simulation, analysis, and robot-aided surgery. Surgical video generation can provide a valuable solution to augment and enrich the real data, and to enable the exploration and evaluation of different surgical scenarios and techniques.
Video generation \cite{Pan_2019_CVPR} is a challenging task that aims to synthesize realistic and coherent video sequences from a given input, such as a text \cite{hu2022mage}, an image \cite{ni2023conditional}, flow maps \cite{Wu2020Future}, or a video \cite{tang2023anytoany}. Video generation has been extensively studied in the computer vision and machine learning literature, and various methods have been proposed, such as  generative adversarial networks (GANs) \cite{lemoing2023waldo} and autoregressive models \cite{weissenborn2019scaling}. Bar \etal \cite{bar2021actiongraph} propose conditioning video synthesis models on action graphs \cite{garg2021unconditional}, while WALDO \cite{lemoing2023waldo} introduces a framework for conditioning the model on segmentation \cite{yeganeh2023scope,yeganeh2023transformers} and flow maps \cite{gao2018im2flow}. However, most of these methods suffer from limitations, such as mode collapse, blurriness, artifacts, and temporal inconsistency \cite{yan2022temporally}. Moreover, most of these methods are not suitable for laparoscopic video generation, as they do not incorporate domain-specific knowledge and constraints, such as anatomy, surgical actions, and camera motion.

In this work, we introduce the novel task of future video generation in laparoscopic surgery, which aims to generate realistic and diverse videos of laparoscopic procedures conditioned on a single frame and an action triplet, defining the action to be performed. For example, given a frame of a cholecystectomy (gallbladder removal) and a triplet of "cut," "cystic duct," and "clip," the task is to generate a video of the surgeon cutting the cystic duct and applying a clip. This task is challenging, as it requires modeling the complex dynamics and interactions of the surgical scene, as well as generating high-fidelity and temporally coherent video frames.

\begin{figure}[tb]
    \centering
    \includegraphics[width=\textwidth]{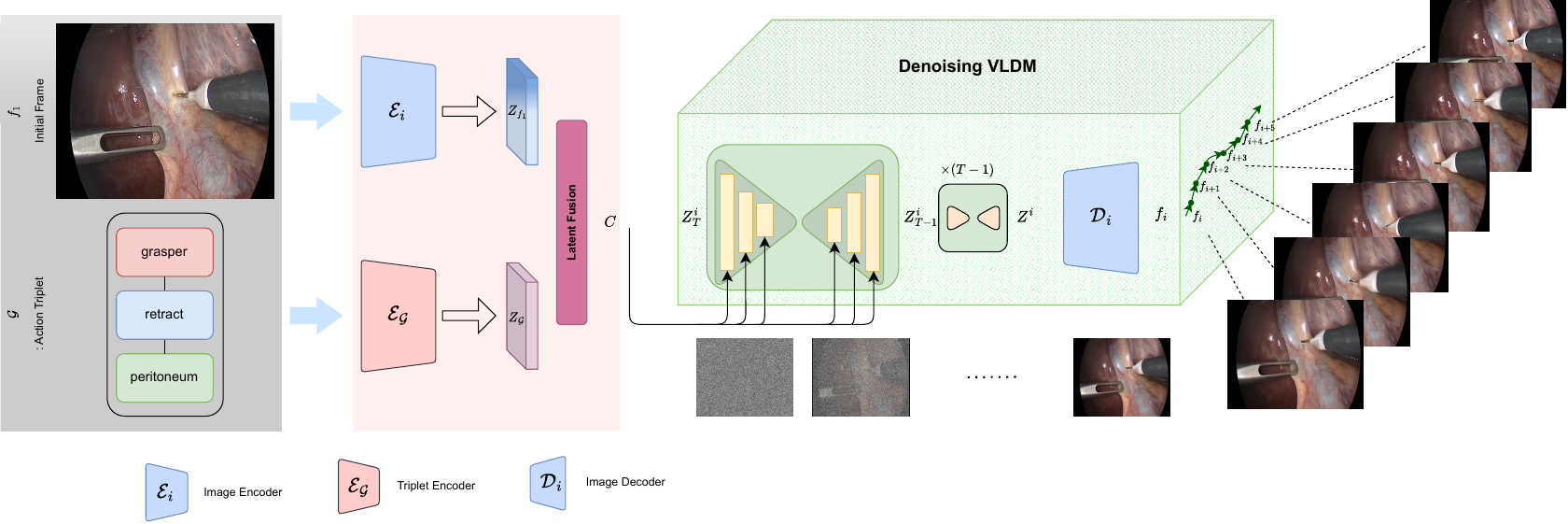}
    \caption{\textbf{VISAGE pipeline.} The Video Latent Diffusion Model (VLDM) generates the future frames conditioned on the fused conditional features obtained from the input frame and the action graph.}
    \label{fig:enter-label}
\end{figure}

To address this task, we propose \methodName{} (VIdeo Synthesis using Action Graphs for Surgery), which leverages the power of action scene graphs and diffusion models. An action scene graph is a structured representation of the organs, the surgical tools, and their relations, which capture the sequential nature of the procedures \cite{nwoye2020recognition}. Image generation \cite{johnson2018image,venkatesh2024exploring,chen2018surgical,marzullo2021towards,mohamadipanah2023generating,farshad2021_MIGS,farshad2023scenegenie} and manipulation \cite{dhamo2020semantic,Farshad_2022_BMVC,jahoda2023prism} models conditioned on action scene graphs have been previously explored and showed promising outcomes in computer vision. Diffusion models are a class of generative models that can produce high-fidelity images and videos by reversing a Markov chain of Gaussian noise \cite{ho2020denoising,rombach2022high,lupke2024physics}. Recent advances in generative AI \cite{yeganeh2022shape,astaraki2023autopaint,bahrami2022intelligent} have led to promising models in unconditional video synthesis in the field of computer vision \cite{blattmann2023align,blattmann2023stable}. By incorporating domain-specific knowledge through the action graph, \methodName{} ensures the generated videos adhere to the expected visual and motion patterns observed in real laparoscopic procedures. Moreover, by using diffusion models, VISAGE avoids the common pitfalls of GANs, such as mode collapse and instability, and achieves state-of-the-art sample quality on the CholecT50 dataset \cite{nwoye2021rendezvous}. To summarize, our main contributions are: (1) We introduce the novel task of future video generation in laparoscopic surgery, which has many potential applications and challenges. (2) We propose \methodName{}, which conditions the diffusion model on action scene graphs for laparoscopic video generation. (3) We conduct extensive experiments on the CholecT50 dataset, and we demonstrate the effectiveness of our method both quantitatively and qualitatively, (4) We compare our method with several baselines and ablations and show it outperforms them in various metrics and qualitative evaluations.

\section{Method}
\label{sec:method}
In this section, we describe the video generation pipeline and the conditioning of the diffusion model.

\vspace{4pt}\noindent\textbf{Definitions} We have a dataset $p_{data}$ of videos and their corresponding action triplets $(s,p,o) \in \mathcal{G}$, such that  $x,\mathcal{G} \sim p_{data}$, where $x \in \mathcal{R}^{K \times 3 \times H \times W}$ is a sequence of $K$ frames in RGB format, with height and width $H$ and $W$, and $\mathcal{G}$ is the set of sequential actions performed in the video defined by the $(s,p,o)$ which are the \textit{subject, object}, and \textit{predicate}, respectively. The goal of the model is to generate the subsequent video frames $x_1,...,x_K$ given $\mathcal{G}$ and $x_0$. The diffusion model utilized for video generation is denoted by $g$ and parameterized by $\theta$. The diffusion model comprises an image encoder and an action graph encoder denoted by $\mathcal{E}_I, \mathcal{E}_\mathcal{G}$, and a decoder denoted by $\mathcal{D}$.

\subsection{Video Latent Diffusion Model}

Latent Diffusion Models (LDMs) \cite{rombach2022high} are generative models based diffusion process. The process comprises a forward (noising) process and a reverse (denoising) process. In the forward process, a data sample $x_0$ is encoded into a latent representation $z_0$ using an encoder $\mathcal{E}$. Then Gaussian noise is gradually added over a series of steps, resulting in a sequence of noisier latent variables until a pure noise latent variable $z_T$ is reached:

\begin{equation}
z_t = \sqrt{1 - \beta_t} z_{t-1} + \sqrt{\beta_t} \epsilon_t, \quad \epsilon_t \sim \mathcal{N}(0, I)
\end{equation}

where $\beta_t$ is the noise schedule for timestep $t$. The reverse process starts from the noise latent variable $z_T$ and iteratively denoises it to recover the original latent representation $z_0$. A neural network $p_\theta$  is used to predict the noise added at each step and subtract it, refining the latent variable at each timestep:

\begin{equation}
z_{t-1} = \frac{1}{\sqrt{1 - \beta_t}} \left( z_t - \frac{\beta_t}{\sqrt{1 - \alpha_t^2}} p_\theta(z_t, t) \right)
\end{equation}

where $p_\theta$ is trained to predict the noise $\epsilon_t$, and $\alpha_t$ is a variance schedule. To extend the diffusion model to video data, we follow the same procedure as \cite{blattmann2023align} and employ 3D convolutional layers, which can capture both spatial and temporal features of the video frames. Similar to \cite{blattmann2023align}, a temporal layer is inserted into the LDM \cite{rombach2022high} after each spatial layer.

\subsection{Action Conditioned Video Generation}
An action scene graph consists of nodes and edges, where each node represents an organ or a surgical tool, and each edge represents the performed action. 
To condition the diffusion model on the action scene graph, we use the graph encoder $\mathcal{E}_\mathcal{G}$ to encode the action scene graph into a latent vector $h$, which is then fused with the encoded initial frame from $\mathcal{E}_I$. The diffusion model is trained on a denoising objective of the form.
\begin{equation}
    \mathbb{E}_{x,h,\boldsymbol{\epsilon},t}{w_t \|\hat{x}_\theta(\alpha_t x + \sigma_t \epsilon, \mathcal{G}) - x \|^2_2}
\end{equation}
, where $(x, \mathcal{G})$ are data-conditioning pairs, $t \sim \mathcal{U}([0, 1])$, $\epsilon \sim \mathcal{N}(0, I)$, and $\alpha_t, \sigma_t, w_t$ are functions of $t$ that influence sample quality.

\vspace{4pt}\noindent\textbf{Conditioning} The diffusion model is conditioned on the action graph using the extracted feature embeddings $h$ from $\mathcal{E}_\mathcal{G}$. We propose and evaluate two types of feature encoding for the graph. The first variation encodes the action graph using a learnable embedding layer, where the embeddings are learned implicitly through optimizing the video generation process. The second variation employs CLIP \cite{radford2021learning} text embeddings for the organ, the tool, and the action. Additionally, we fine-tune the CLIP text encoder using the image generation objective.

\vspace{4pt}\noindent\textbf{Feature Fusion} We propose two variations of feature fusion between the graph and image embeddings. The first variation introduces a linear layer that combines the latent features through concatenation. The second variant employs a cross-attention layer for the future fusion of the action graph conditioning and the image features.

\vspace{4pt}\noindent\textbf{Sampling Process}
To generate a video conditioned on the frame $x_0$ and the triplet $(s,p,o)$, we first encode the action scene graph into a latent vector $h = \mathcal{E}_\mathcal{G}(s,p,o)$. We then initialize the noise sample $z_T$ by adding Gaussian noise to the single frame and then apply the reverse process of the diffusion model to denoise $z_T$ into $z_0$. At each timestep $t$, we use the fused latent vector $z_t$ to condition the network $g_\theta$ on the action scene graph and the initial frame.

\subsection{Data Preprocessing} 
The videos in CholecT50 \cite{nwoye2021rendezvous} dataset consist of an average of 1.7K frames, which is larger than the number of frames our baseline models can process. Therefore, we need to divide the videos into \textit{scenes}, which are smaller clips of the videos in the dataset. To be able to generate temporally consistent videos, generative video models are better trained on datasets that consist of videos that are also temporally consistent \cite{blattmann2023stable}. However, raw videos are prone to containing motion inconsistencies, such as jumps between scenes, which we will call \textit{scene cuts}, and still videos with very little motion as \textit{static scenes}. When trained on static scenes, the model might learn to predict little to no motion, or when trained on videos with scene cuts, the model might learn to predict jumps between frames. Therefore, we process the dataset before training to mitigate possible issues. Due to SVD's \cite{blattmann2023stable} limitation for the maximum number of frames to process for each video, which is 7, we fix 7 as the frame count of each scene. We obtain 7-framed scenes after applying the methods mentioned in the following sections, which filter the videos to eliminate scene cuts and static scenes.

\vspace{4pt}\noindent\textbf{Scene Cut Detection}
To detect scene cuts, which are sudden changes in a sequence of images, we first use the PySceneDetect\footnote{https://github.com/Breakthrough/PySceneDetect} library. PySceneDetect analyzes videos by comparing the neighboring frames for changes in intensity, brightness, and HSV color space between frames and can also look for fast camera movements. The threshold for the sensitivity in detection can be defined, which we set to 0.27 as suggested. It returns a list of frame IDs where a cut is detected, which is our rough start to divide a video. After the first division by PySceneDetect, we further divide the detected scenes into 7-framed portions, where we add another rule regarding the triplet annotations. We force that all frames in the 7-framed scene at least have one common triplet. This allows us to use the triplet information also as a cut detection method since a frame with no common triplet with other frames in the scene would mean a jump in the depicted motion. We also eliminate the scenes that include frames with empty triplet annotation, such as an undefined triplet or undefined elements for the predicate, instrument, or target.

\vspace{4pt}\noindent\textbf{Static Scene Detection}
Static scenes in our videos occur in two ways: Scenes with little motion and scenes that are completely blacked out due to privacy. We detect the latter simply by looking at pixel intensities and eliminating scenes, including all-black frames. They can also be detected by the empty triplet check. The first type of static scene is harder to detect. As in \cite{blattmann2023stable}, we tried to detect the static scenes by using optical flow to calculate a motion score; however, the scores did not represent the motion well enough to be used in detection. Therefore, the static scenes of the first type are detected by the empty triplets or triplet elements; for example, scenes where only the (target) organ is in the frame and no action is performed can be detected by checking the triplet annotations.

\begin{table}[tb]
\centering
 \caption{\textbf{Comparison to SOTA.} Quantitative evaluation of video synthesis on CholecT50 \cite{nwoye2021rendezvous}.}
\begin{tabular}{l@{\hspace{10pt}} c@{\hspace{10pt}} c c c c} 
 \hline
 Model & Input & FVD (↓) & PSNR (↑) & LPIPS (↓) & SSIM (↑) \\
 \hline\hline
 CoDi \cite{tang2023anytoany} & Image + Triplet & 6,944 & 9.8 & 0.82 & 0.31 \\ 
 WALDO \cite{lemoing2023waldo} & Image + Seg. + Flow & 3,413 & 11.6 & 0.72 & 0.34 \\
 LFDM \cite{ni2023conditional} & Image + Triplet  & 1,957 & 12.0 & 0.54 & \textbf{0.71} \\
 SVD \cite{blattmann2023stable} & Image & 3,870 & 14.8 & 0.51 & 0.47 \\
 SVD + FT \cite{blattmann2023stable} & Image & 1,931 & \underline{18.2} & 0.40 & 0.55 \\
 \methodName{}-T (Ours) & Image + Triplet & \textbf{1,780} & 18.1 & 0.39 & \underline{0.56} \\
  \methodName{}-I (Ours) & Image + Triplet & \underline{1,875} & \textbf{18.3} & \textbf{0.38} & \underline{0.56} \\
 \hline
 \end{tabular}
 \label{tab:sota}
\end{table}

\section{Experiments and Results}
\subsection{Experimental Setup} 
\vspace{4pt}\noindent\textbf{Dataset.} For all the experiments, we use the CholecT50 Dataset \cite{nwoye2021rendezvous}, which is a laparoscopic video dataset of 50 videos for recognizing and localizing surgical action triplets. Each action triplet includes a predicate defining the action, an object (surgical instrument), and a target (organ), which are used as annotations of the individual frames in videos. A frame can be annotated by zero or up to 3 triplets. In addition, the surgical phase for each frame is annotated. There are a total of 100 triplets, 10 predicates, 6 instruments, 15 targets, and 7 phases.

\vspace{4pt}\noindent\textbf{Implementation Details.}
All our models were trained for 10,000 iterations using Adam Optimizer and a learning rate of 1e-5. We applied an L2 loss between the latent space of the generated 7-frames and their corresponding ground truth to guide the training process effectively.

\vspace{4pt}\noindent\textbf{Evaluation Metrics.} We employ the common metrics in video generation for computer vision applications to assess the performance of our model and the baselines. Frechet Video Distance (FVD) \cite{unterthiner2018towards} compares the probability distribution of generated videos with real videos. Peak Signal-Noise Ratio (PSNR) measures the fidelity of reconstruction. Learned Perceptual Image Patch Similarity (LPIPS) quantifies the perceived similarity between two videos. Structural Similarity Index Measure (SSIM) measures changes in structural information between original and generated videos.

\begin{table}[tbh]
\centering
 \caption{\textbf{Ablation Study.} Different condition encoding and fusion techniques. \textbf{Att.} stands for cross-attention fusion either with an image or the triplet as the query. \textbf{FT} stands for fine-tuning.}
\begin{tabular}{l@{\hspace{10pt}} c@{\hspace{10pt}} c c c c} 
 \hline
 Conditioning & Fusion & FVD (↓) & PSNR (↑) & LPIPS (↓) & SSIM (↑) \\
 \hline\hline
 - & - & 1,931 & 18.2 & 0.40 & 0.55 \\
 \hline
 Triplets & Linear & 2,461 & 17.0 & 0.47 & \textbf{0.67} \\

 Triplets & Att. + I & \underline{2,299} & \underline{17.4} & \underline{0.45 }& \textbf{0.67} \\
 
 \hline
CLIP & Linear & 2,022 & 17.3 & 0.41 & 0.53 \\
 CLIP + FT & Linear & 1,851 & 17.6 & 0.39 & 0.54 \\ 
  CLIP + FT & Att. + I & 1,875 & \textbf{18.3} & \textbf{0.38} & \underline{0.56} \\
 CLIP + FT &  Att. + T & \textbf{1,780} & 18.1 & 0.39 & \underline{0.56} \\
 \hline
 \end{tabular}
 \label{tab:ablation}
\end{table}
\vspace{4pt}\noindent\textbf{Baselines.}
We evaluated four distinct video generation architectures on the CholecT50 dataset \cite{nwoye2021rendezvous}, specifically WALDO \cite{lemoing2023waldo}, CoDi \cite{tang2023anytoany}, Stable Video Diffusion (SVD) \cite{blattmann2023stable}, and Latent Flow Diffusion Model (LFDM) \cite{ni2023conditional}. The comparison among these models involved conducting inference procedures on each model, followed by a comprehensive assessment of the generated outputs through both qualitative and quantitative analyses. It is important to note that these models were originally trained on natural video datasets; thus, the application of these models to egocentric videos from laparoscopic cholecystectomy represents an out-of-distribution scenario relative to their training data. The selection of WALDO as one of the baseline models stemmed from its unique utilization of both flow maps and segmentation maps, effectively integrating high-level semantic information into the video generation process. CoDi was chosen due to its versatile nature, as it inherently supports various modalities such as text, audio, image, and videos, both as input and output. LFDM was included owing to its capability to synthesize scene flow, particularly for its ability to synthesize scene flow in the process of video generation, given an initial image and an action class label. Lastly, the inclusion of SVD was motivated by its training on rich and diverse datasets, coupled with its inherent capacity to model temporal consistency, making it a significant contender for benchmarking against other models in the context of video generation tasks. 

\subsection{Results}
\noindent\textbf{Quantitative Results}
\autoref{tab:sota} shows the comparison of different SOTA architectures on the surgical video generation task.
As it can be seen, LFDM \cite{ni2023conditional} has the lowest FVD and, on the other hand, the highest SSIM among inference-only models. It also ranks high in PSNR and LPIPS.
\methodName{} achieves the best overall performance across all models except for SSIM. SSIM captures the structural similarity between the frames, while FVD compares the general distribution of the generated data compared to the real distribution. \methodName{} improves the FVD by a large margin compared to the baseline models, demonstrating its ability to adhere to the real data distribution. It is noteworthy that, although more complex GAN-based models such as WALDO receive additional information such as the segmentation and the flow map, they fail in generating realistic samples and are prone to mode collapse. 

\def\mywidth{55pt}
\def\myheight{28pt}
\begin{figure}[tb]
\centering
\includegraphics[width=0.20\textwidth,height=40pt]{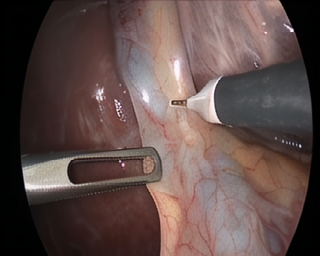} \includegraphics[width=0.2\textwidth]{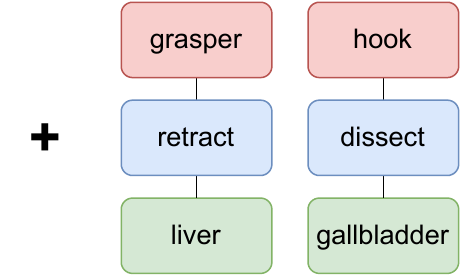}
\begin{tabular}{cc@{}c@{}c@{}c@{}c@{}c}
    & $f_1$ & $f_2$ & $f_3$ & $f_4$ & $f_5$ & $f_6$\\
    \vspace{2 pt}
    \rotatebox[origin=c]{90}{\tiny{GT}} & \raisebox{-.5\height}{\includegraphics[width=\mywidth]{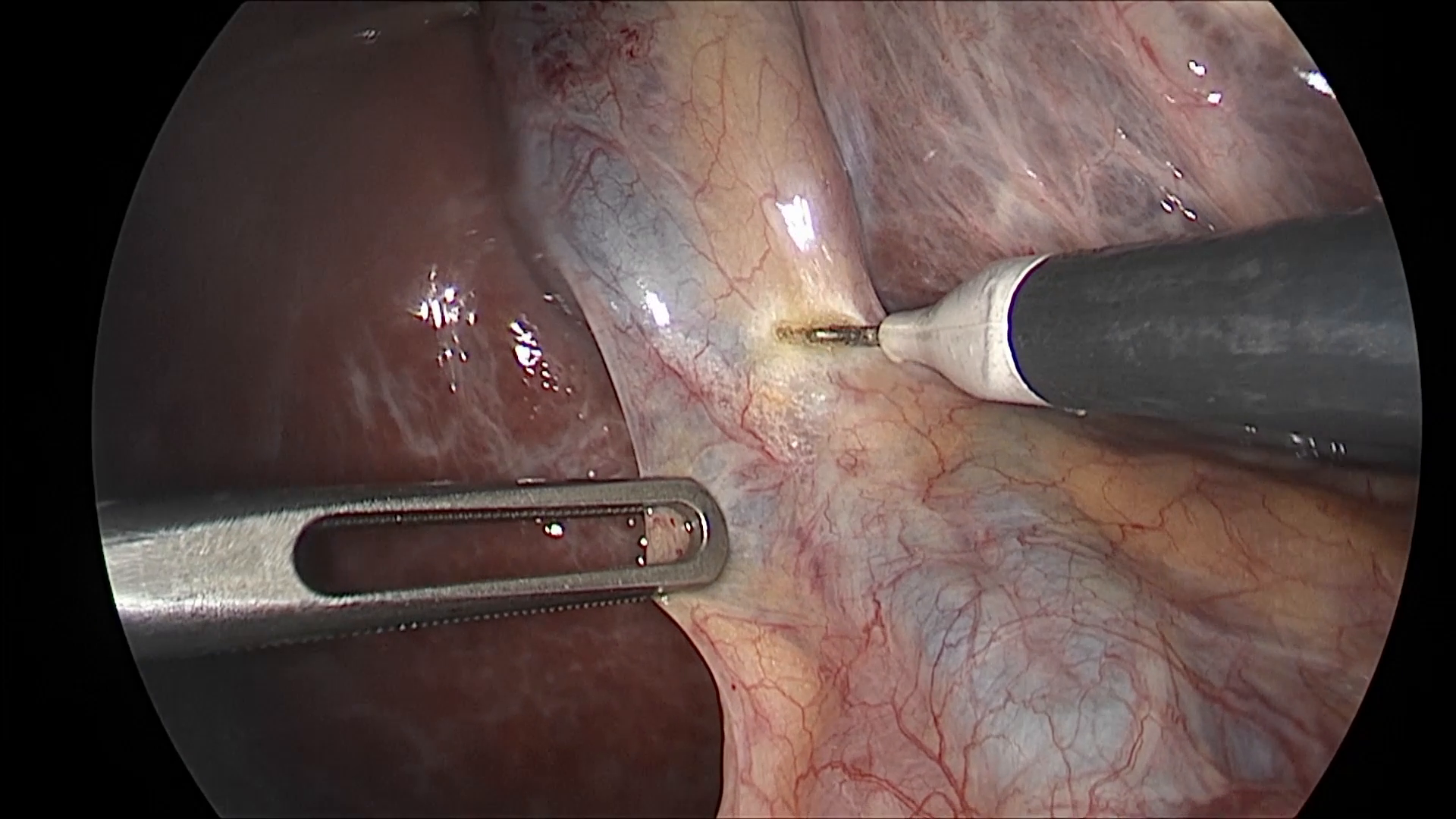}} & \raisebox{-.5\height}{\includegraphics[width=\mywidth]{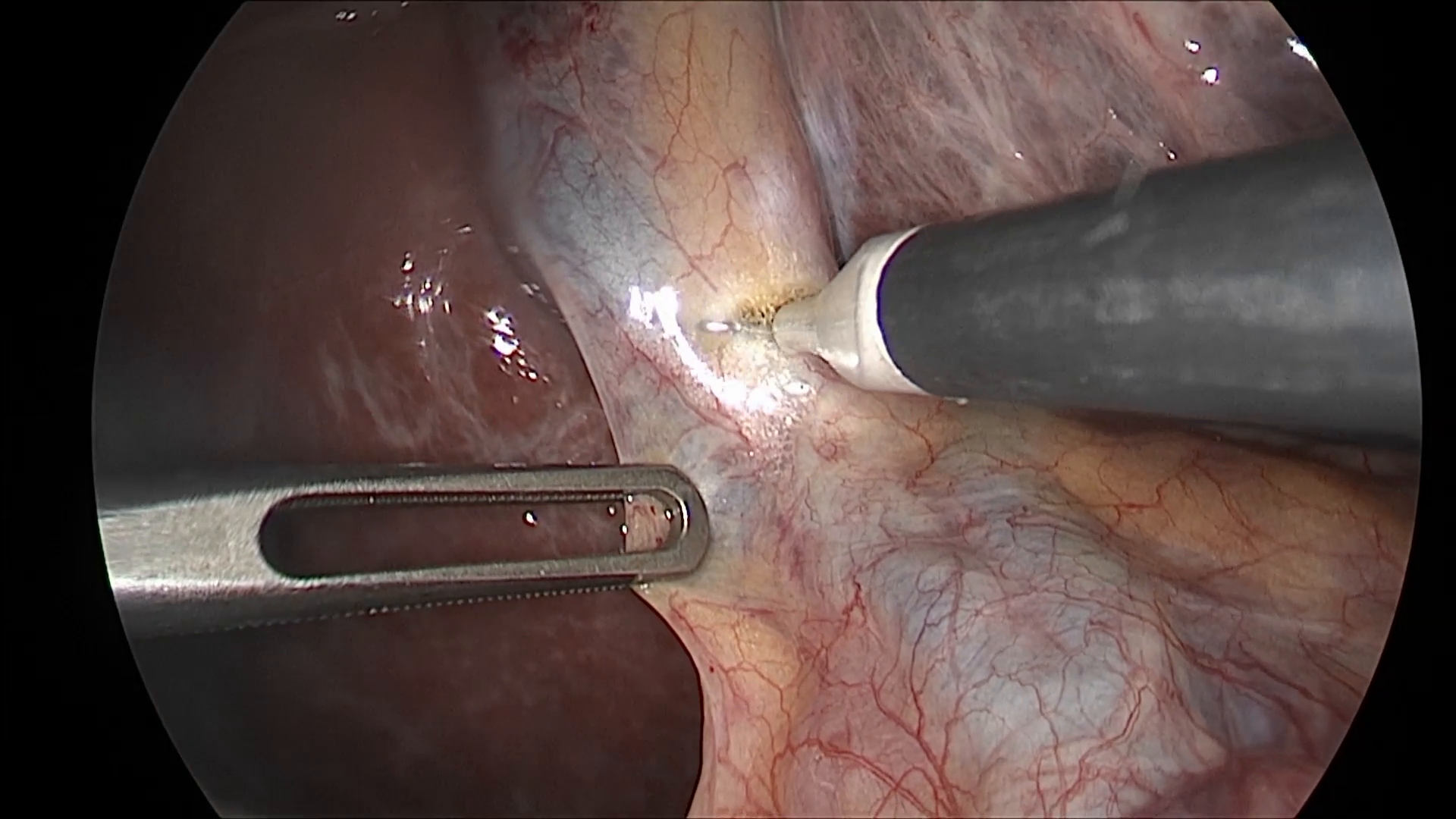}} & \raisebox{-.5\height}{\includegraphics[width=\mywidth]{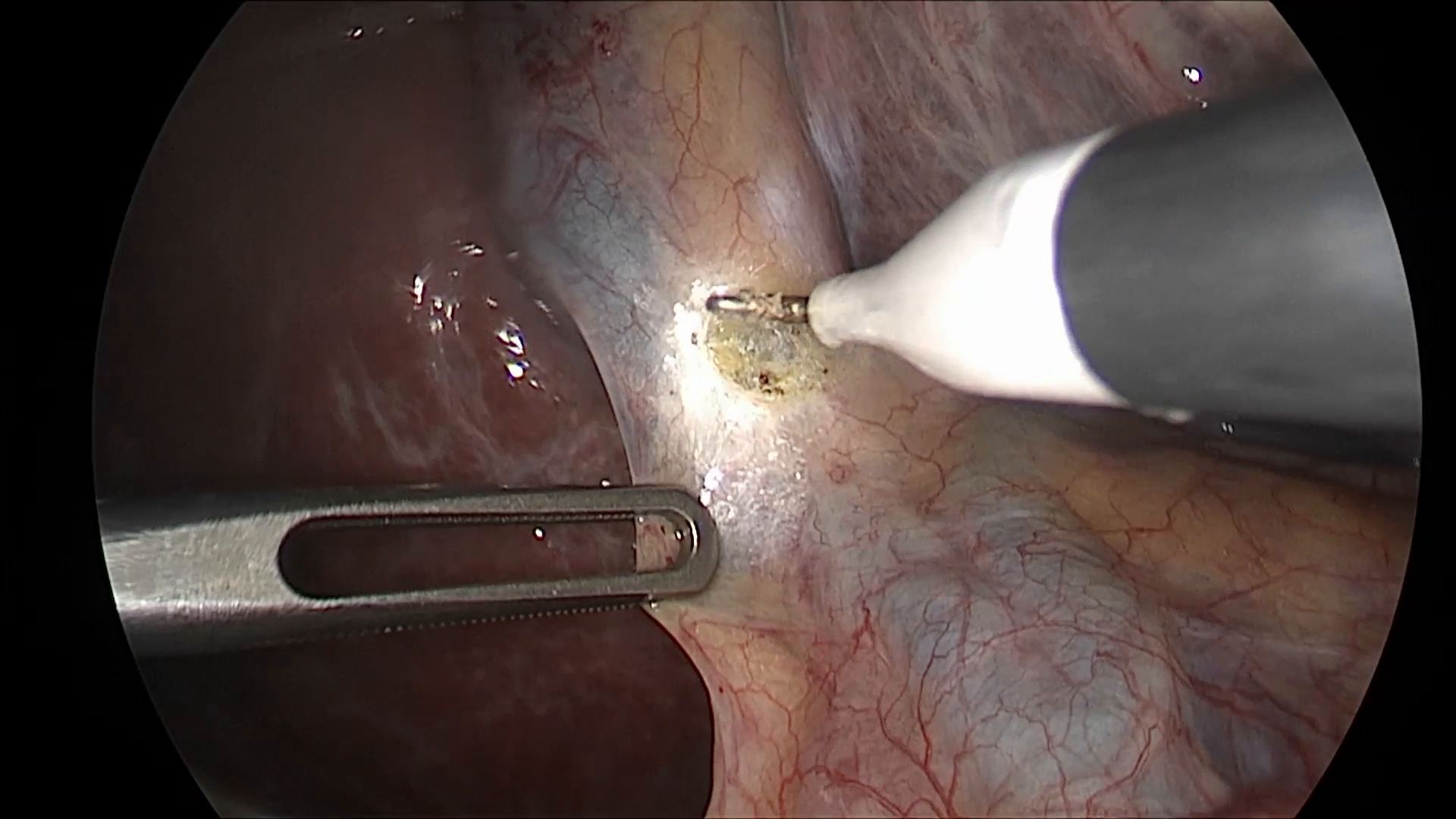}} & \raisebox{-.5\height}{\includegraphics[width=\mywidth]{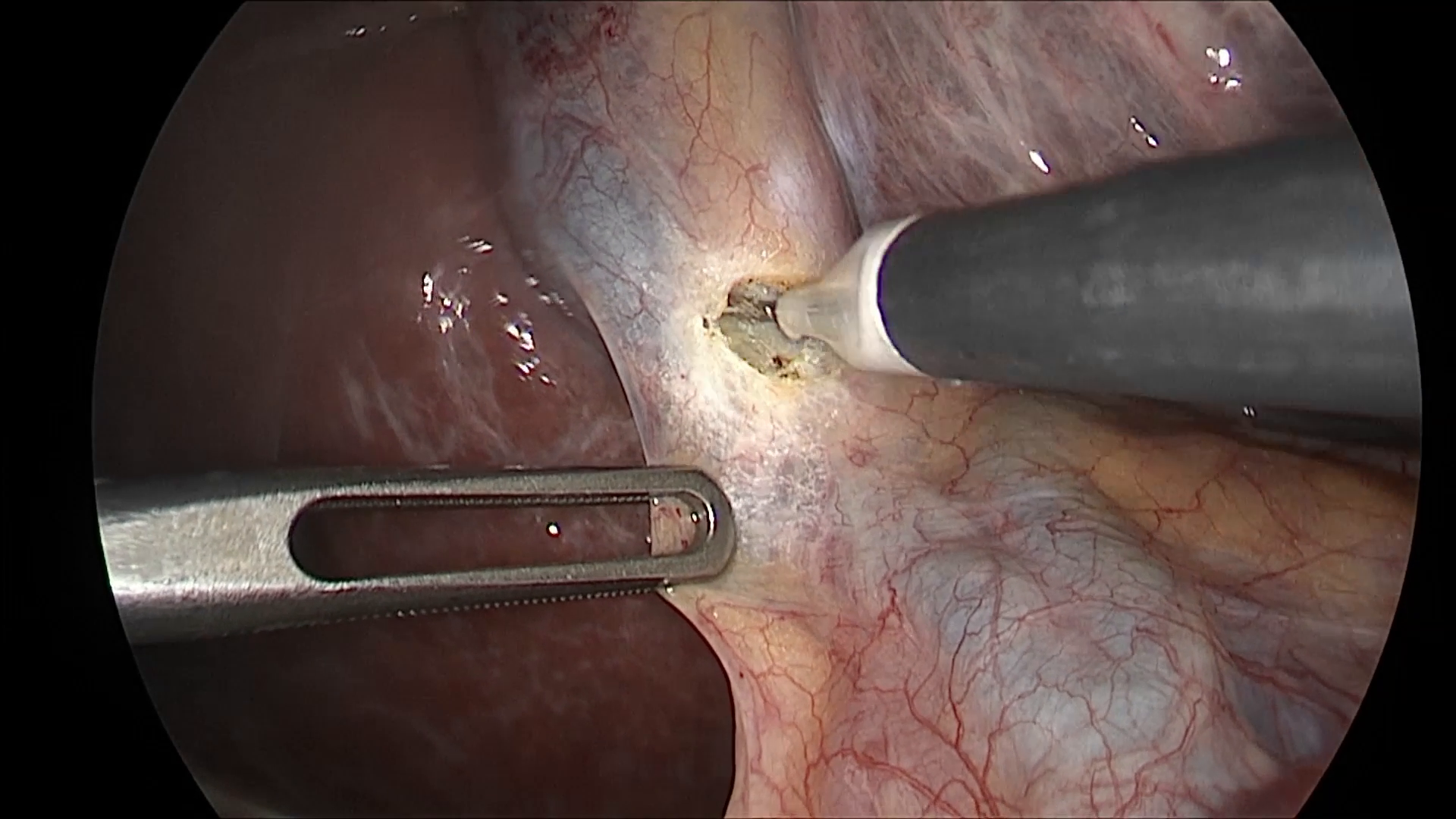}}  & \raisebox{-.5\height}{\includegraphics[width=\mywidth]{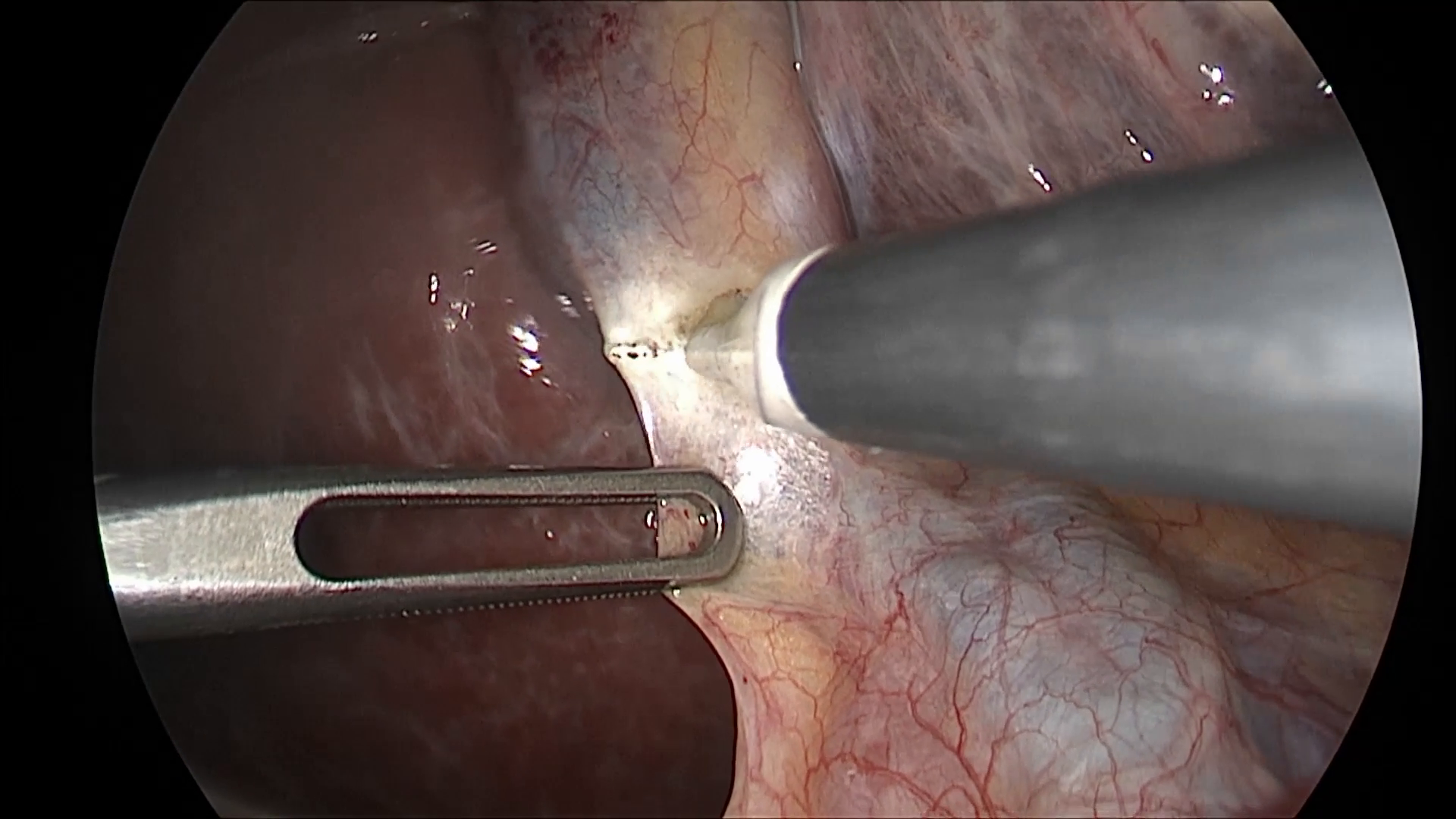}} & \raisebox{-.5\height}{\includegraphics[width=\mywidth]{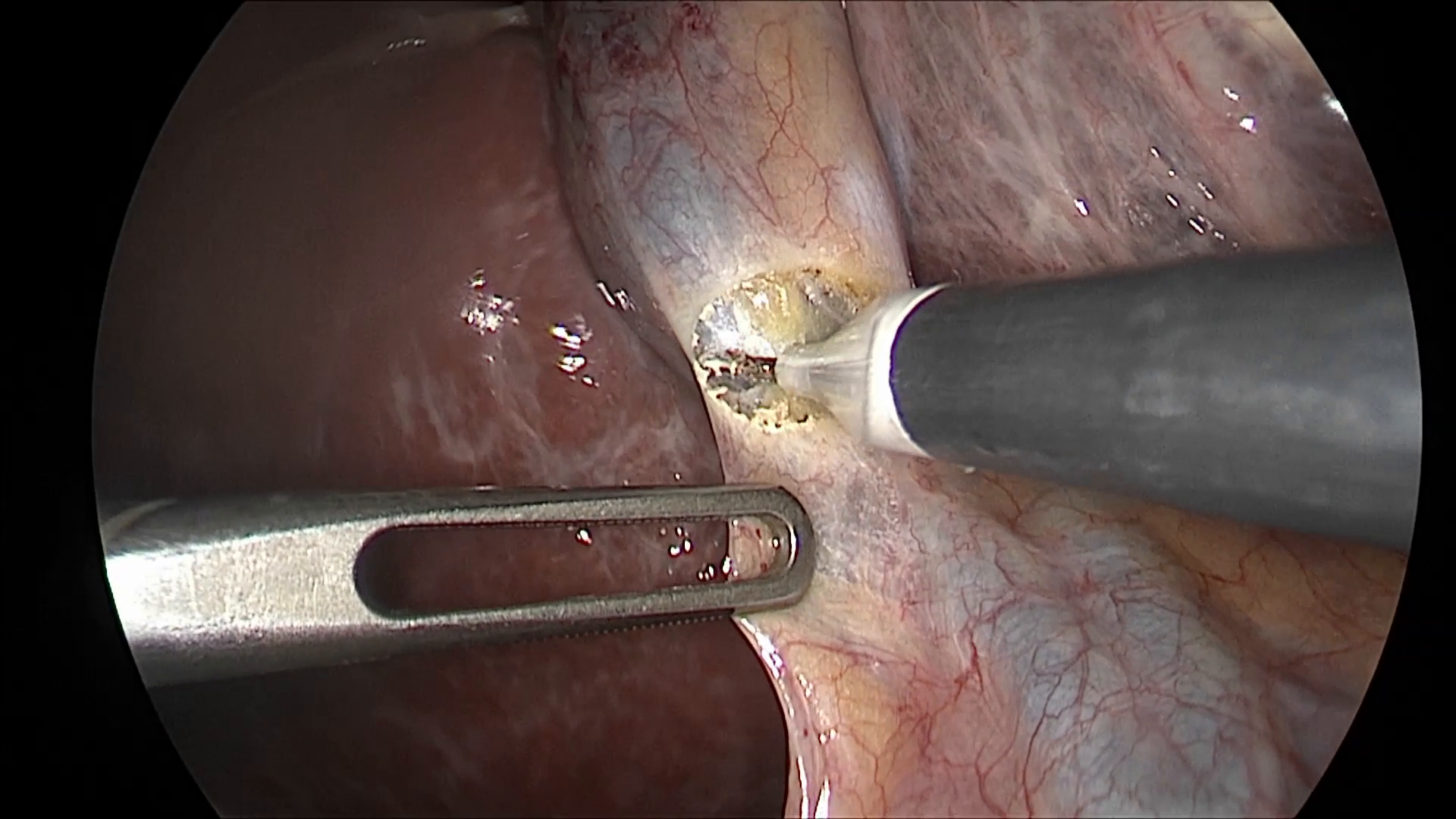}}  \\
     \vspace{2 pt}
     \rotatebox[origin=c]{90}{\tiny{CoDi}} & \raisebox{-.5\height}{\includegraphics[width=\mywidth]{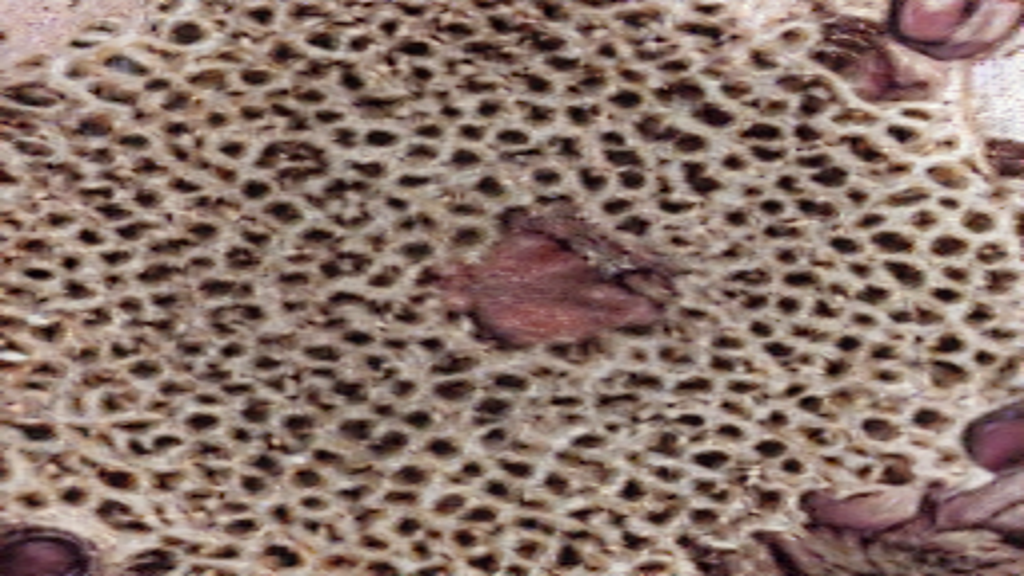}} & \raisebox{-.5\height}{\includegraphics[width=\mywidth]{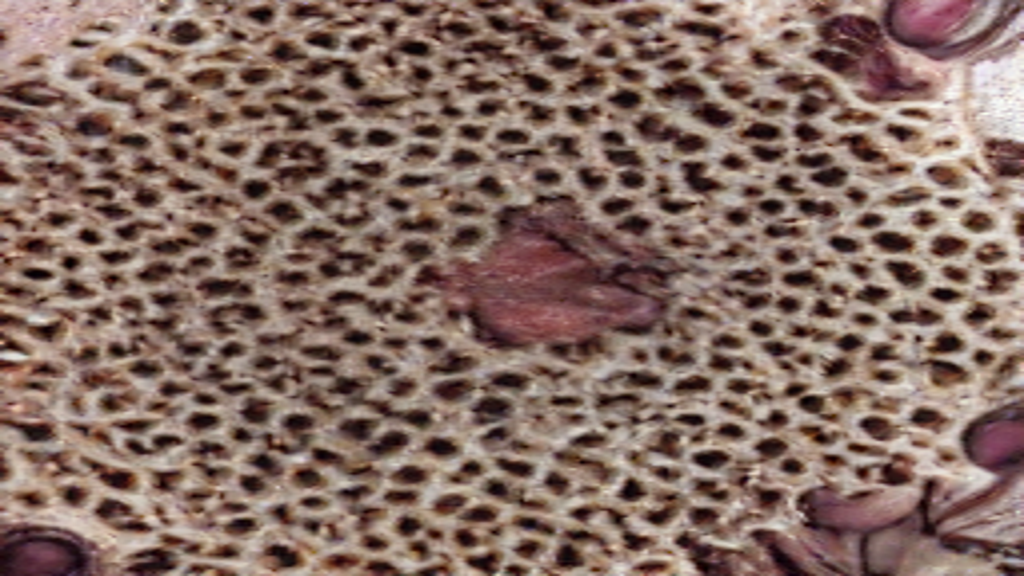}} & \raisebox{-.5\height}{\includegraphics[width=\mywidth]{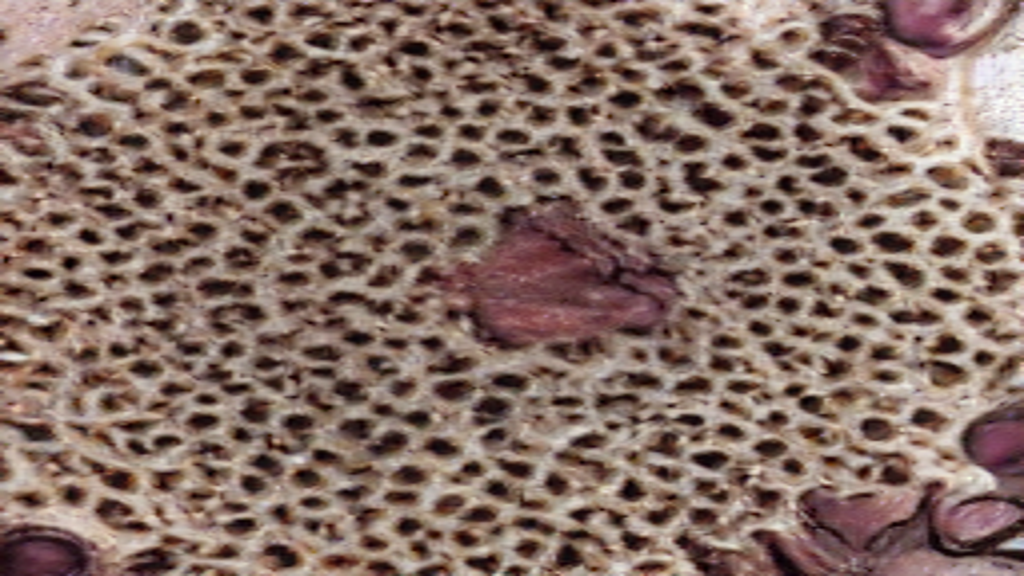}} & \raisebox{-.5\height}{\includegraphics[width=\mywidth]{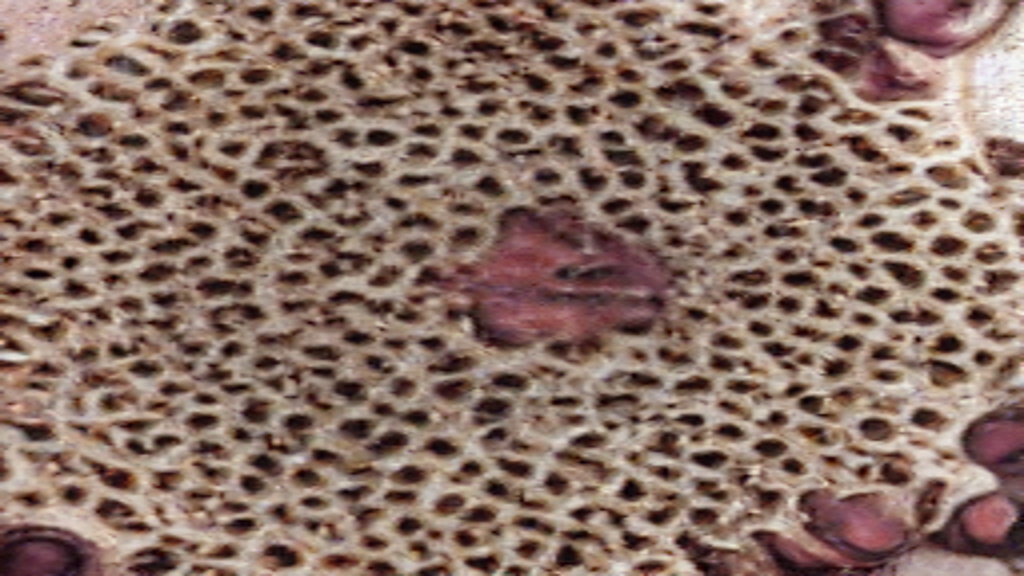}}  & \raisebox{-.5\height}{\includegraphics[width=\mywidth]{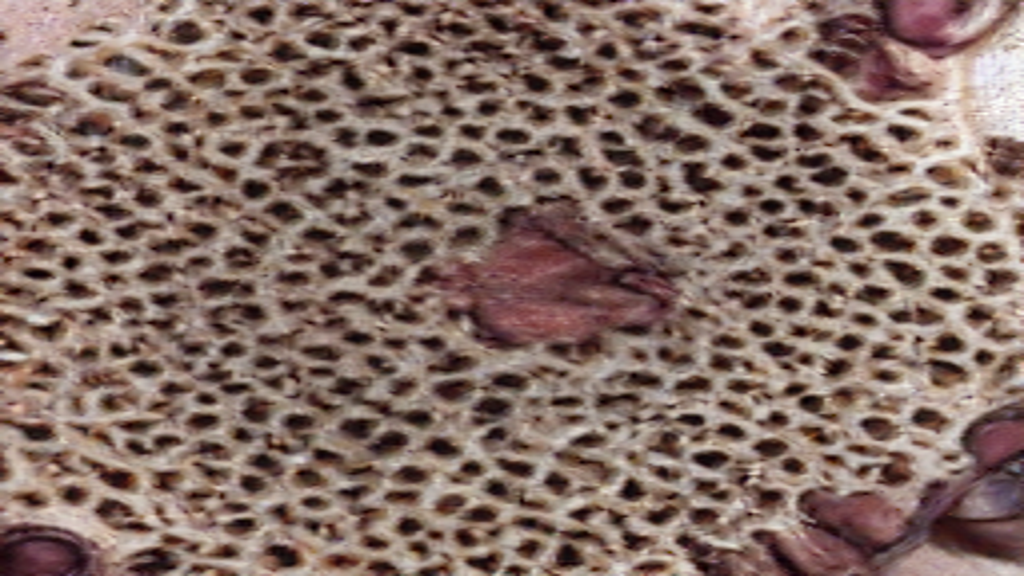}}  & \raisebox{-.5\height}{\includegraphics[width=\mywidth]{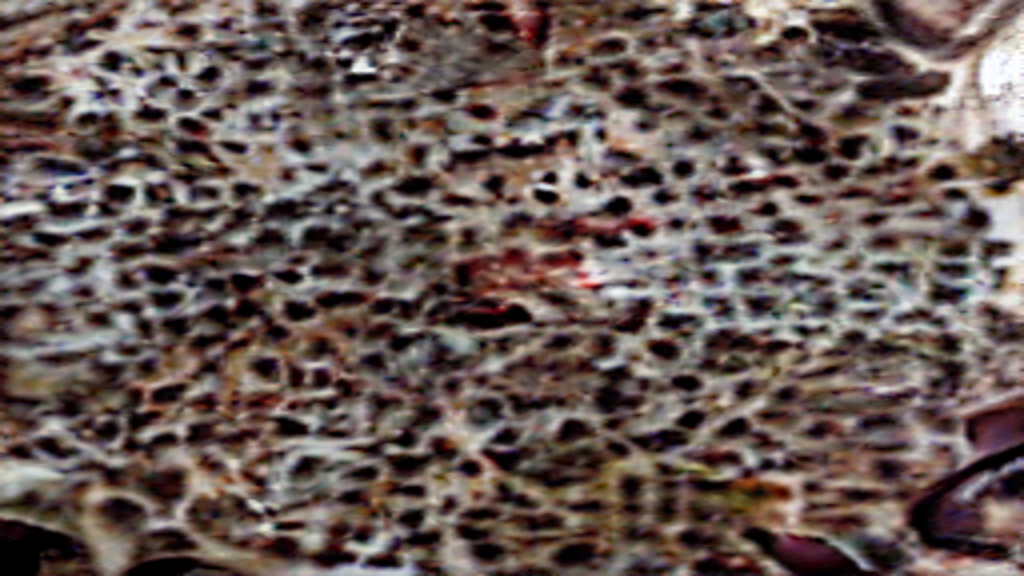}} \\
    \vspace{2 pt}
     \rotatebox[origin=c]{90}{\tiny{Waldo}} & \raisebox{-.5\height}{\includegraphics[width=\mywidth]{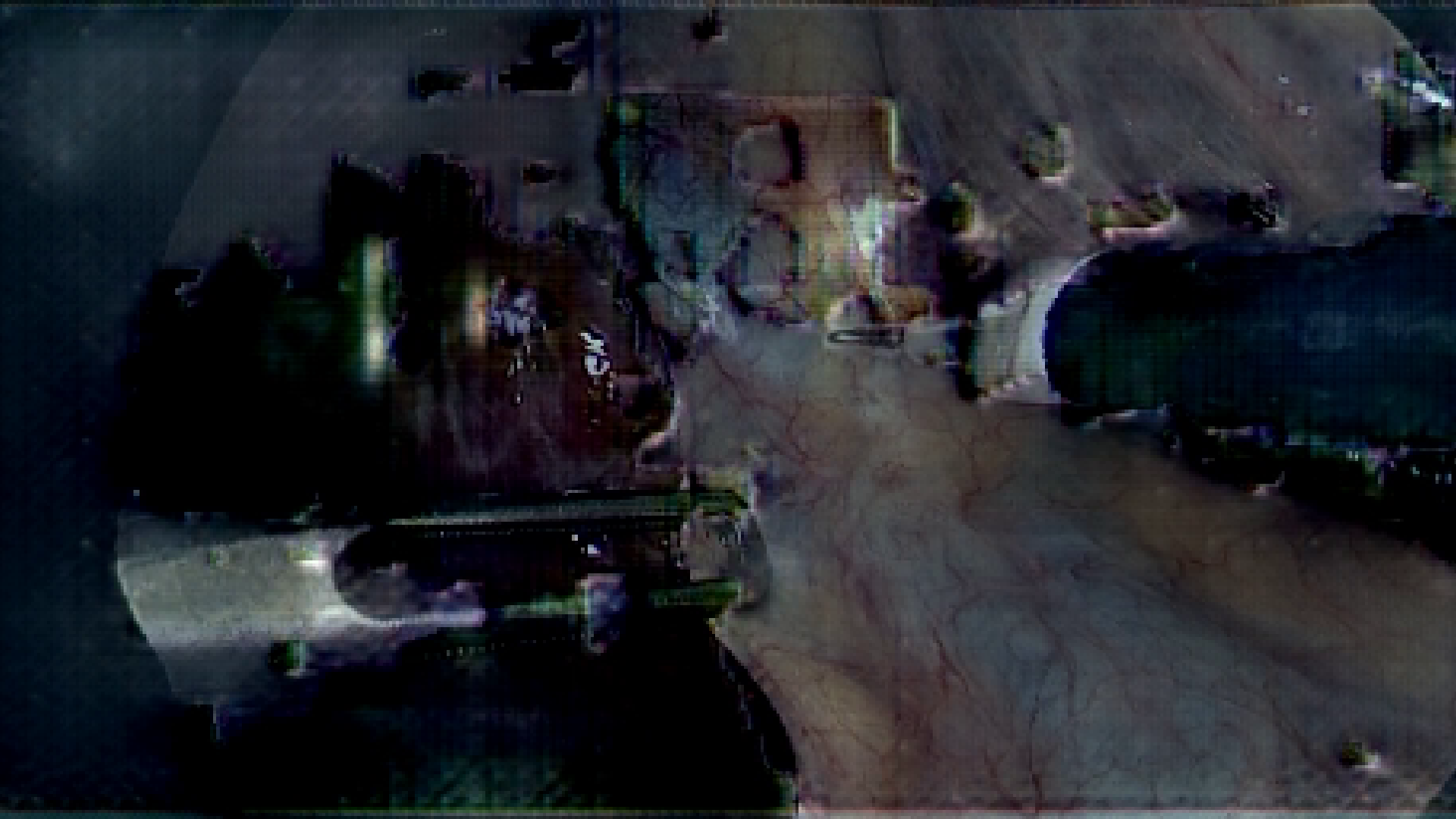}} & \raisebox{-.5\height}{\includegraphics[width=\mywidth]{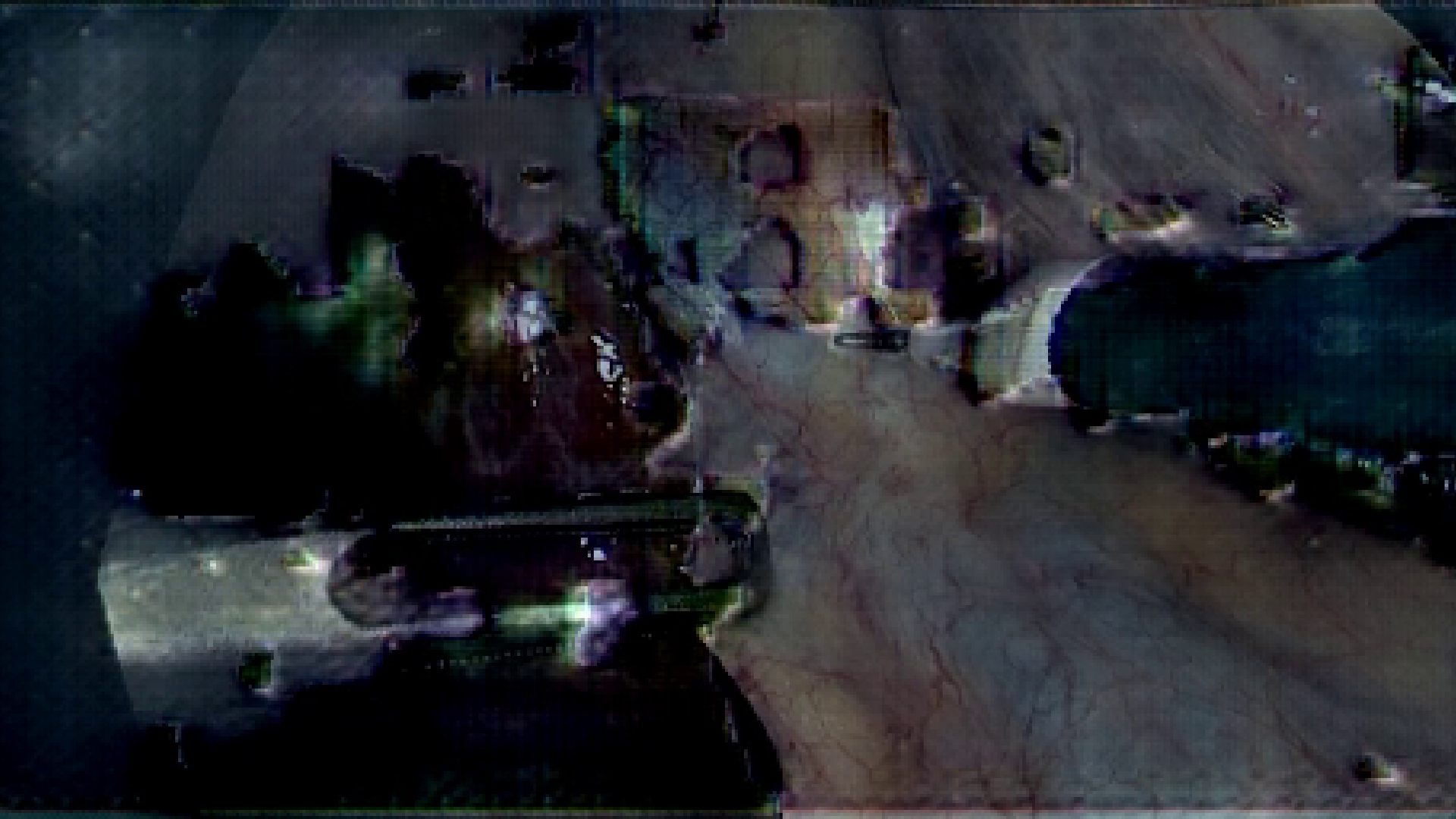}} & \raisebox{-.5\height}{\includegraphics[width=\mywidth]{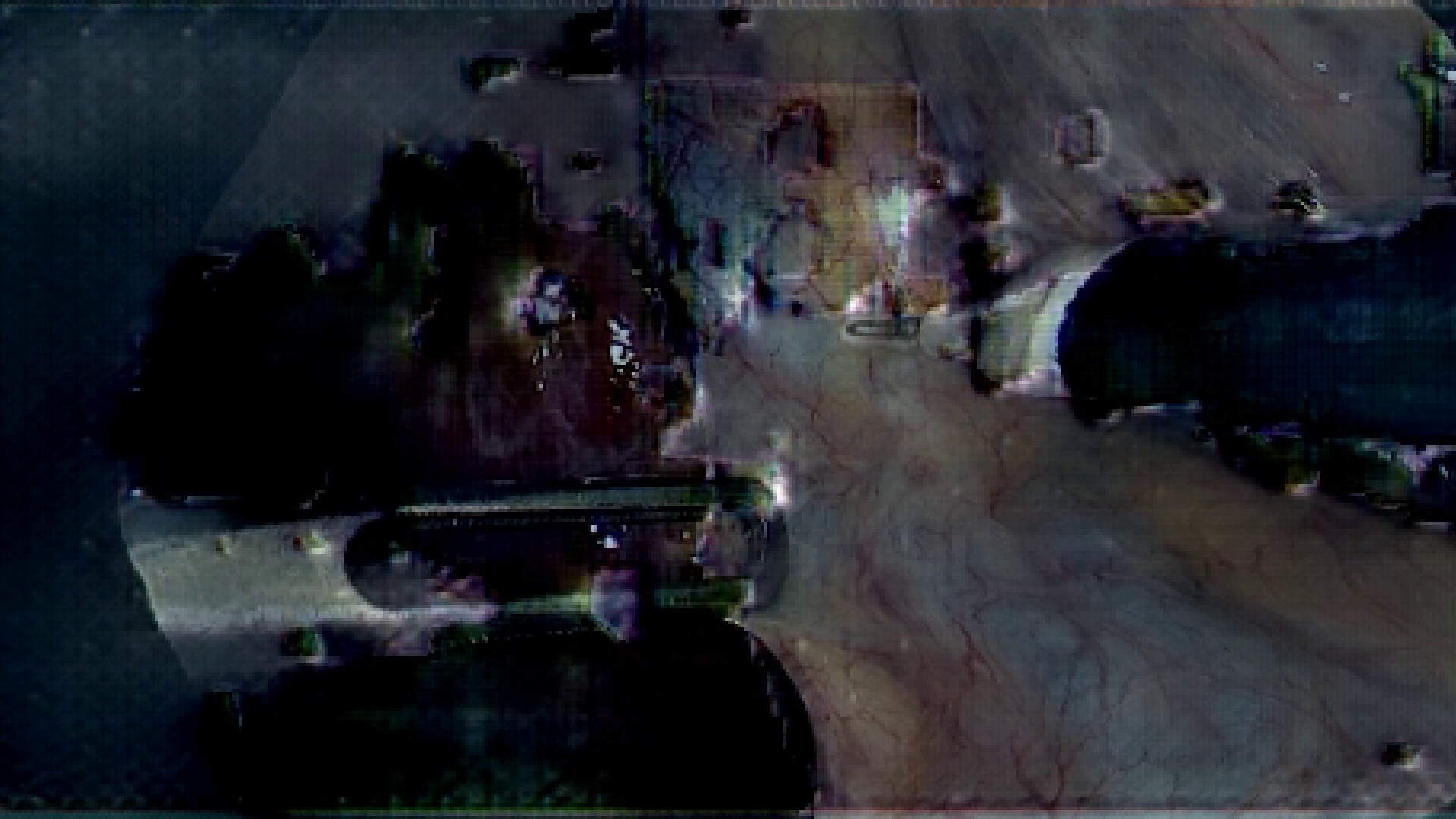}} & \raisebox{-.5\height}{\includegraphics[width=\mywidth]{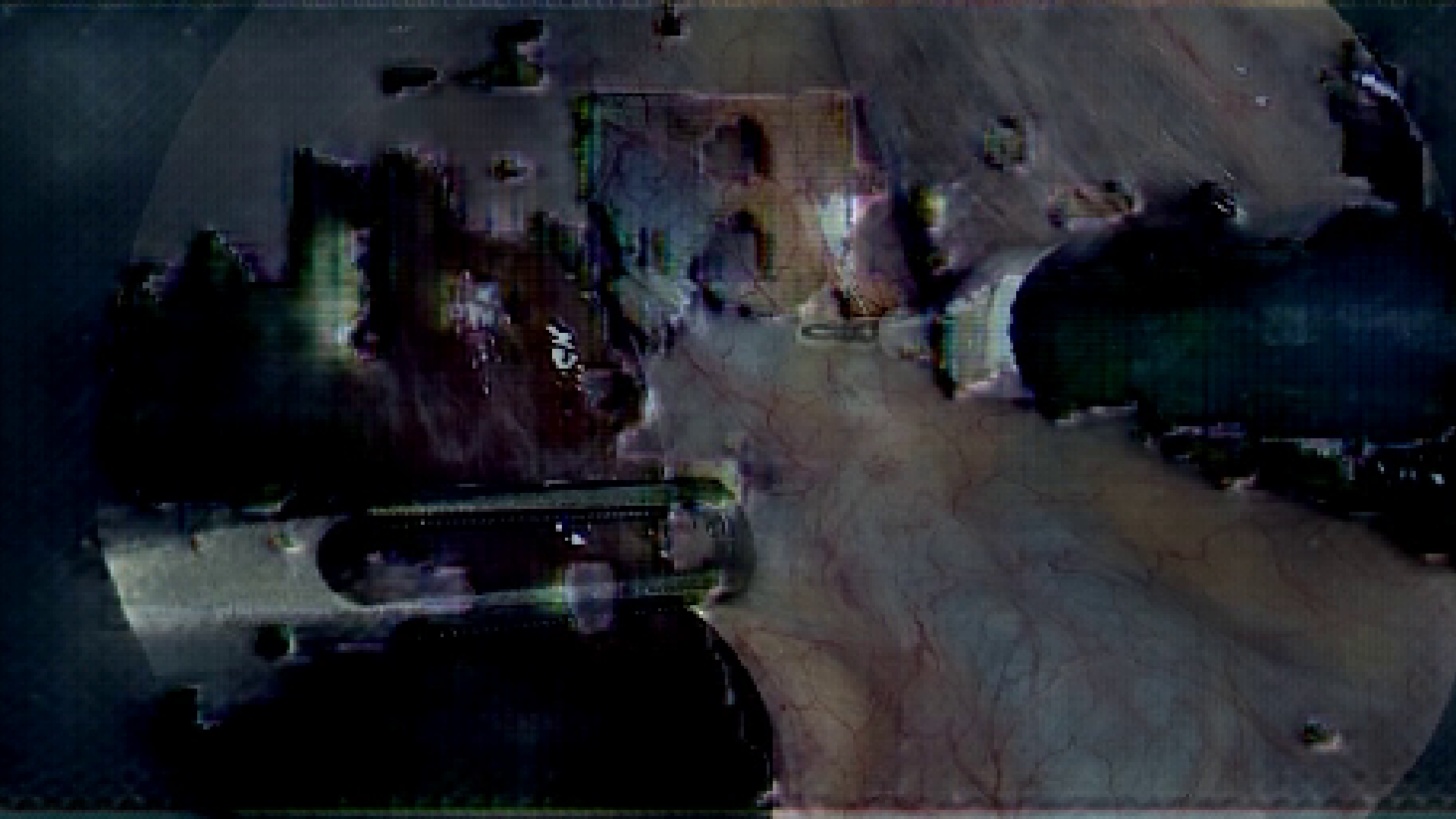}}  & \raisebox{-.5\height}{\includegraphics[width=\mywidth]{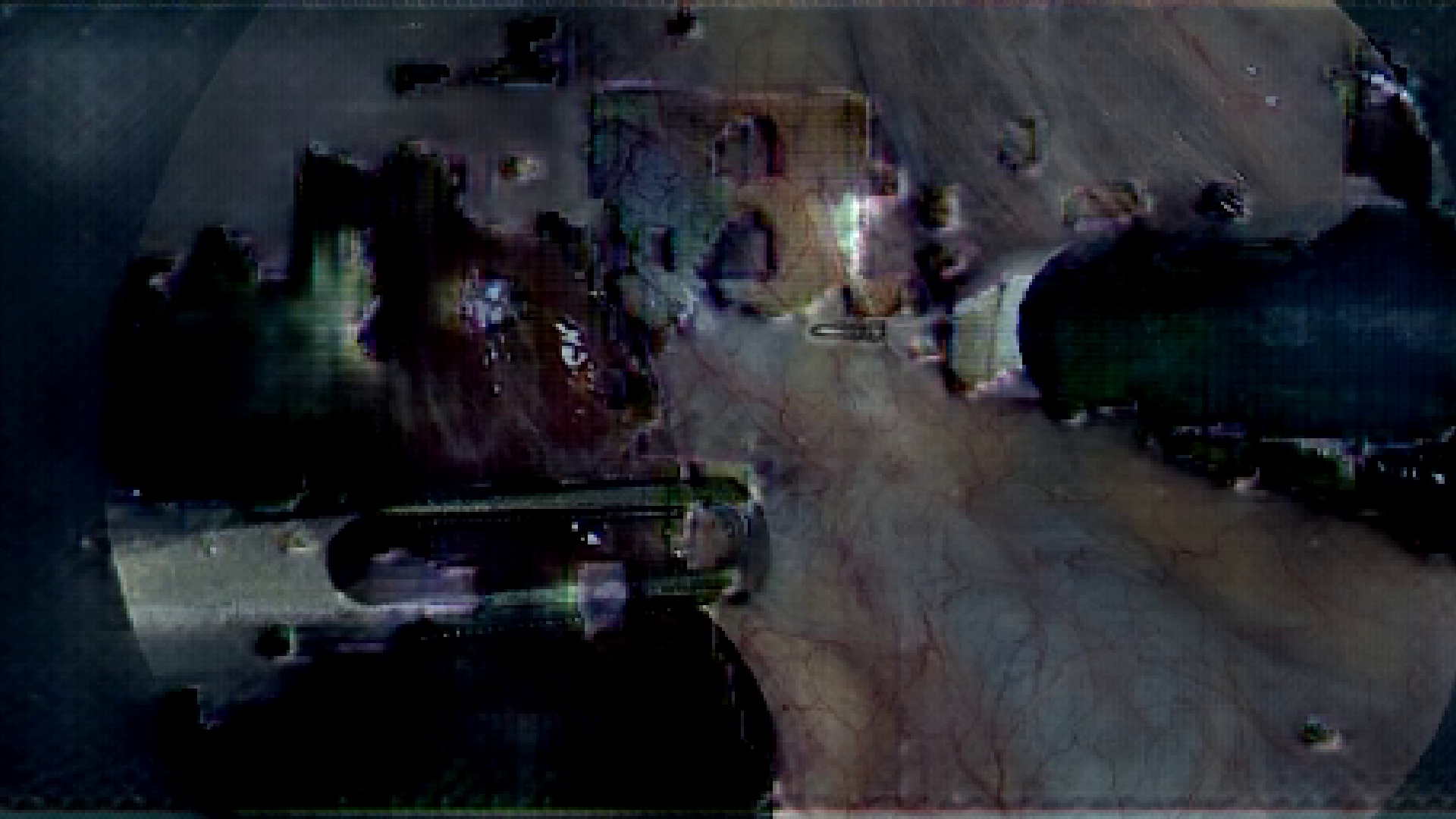}} & \raisebox{-.5\height}{\includegraphics[width=\mywidth]{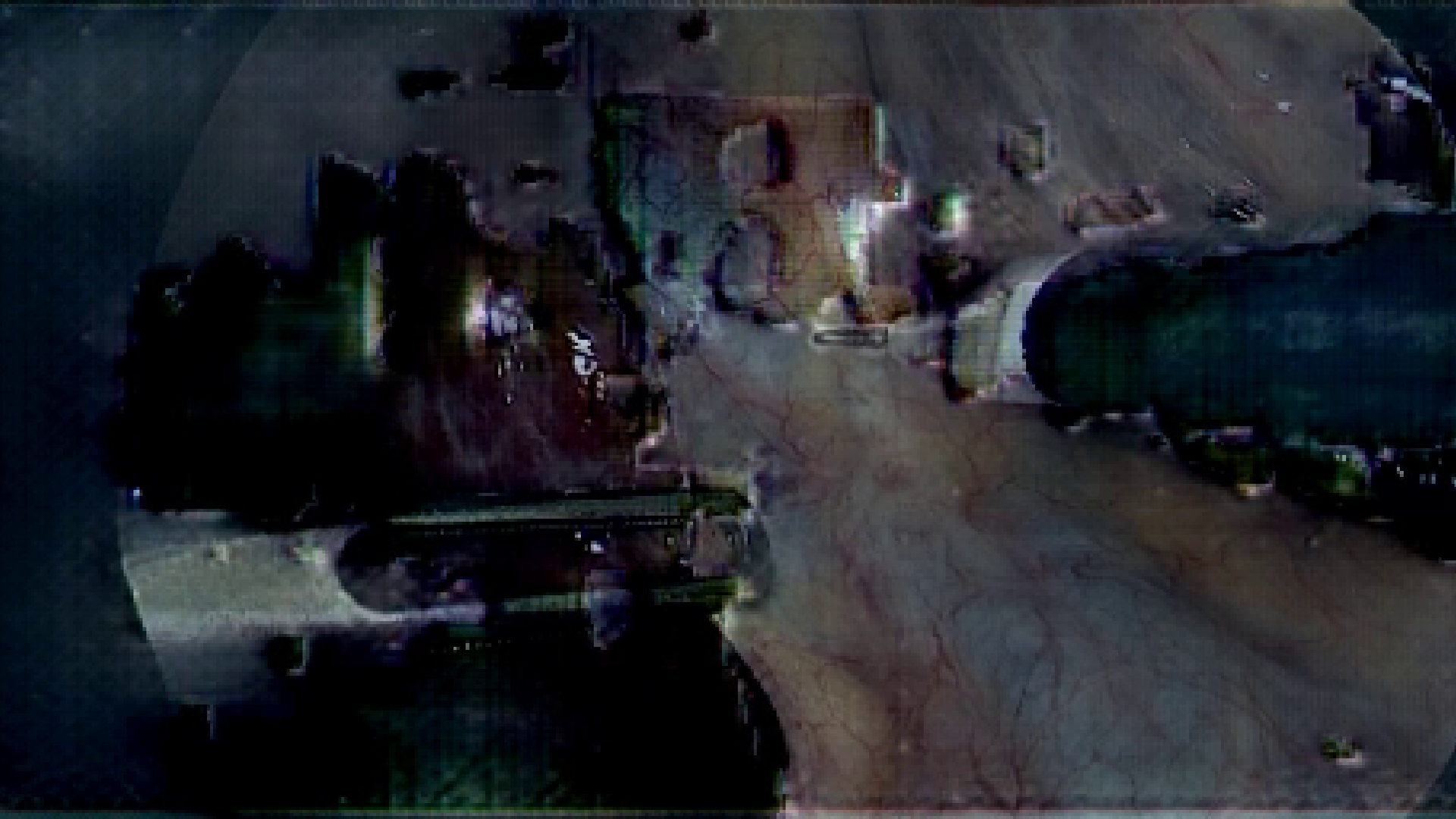}}\\
     \vspace{2 pt}
     \rotatebox[origin=c]{90}{\tiny{LFDM}} & \raisebox{-.5\height}{\includegraphics[width=\mywidth]{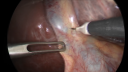}} & \raisebox{-.5\height}{\includegraphics[width=\mywidth]{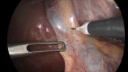}} & \raisebox{-.5\height}{\includegraphics[width=\mywidth]{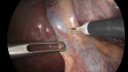}} & \raisebox{-.5\height}{\includegraphics[width=\mywidth]{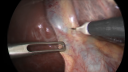}}  & \raisebox{-.5\height}{\includegraphics[width=\mywidth]{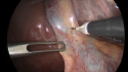}}  & \raisebox{-.5\height}{\includegraphics[width=\mywidth]{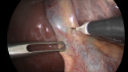}} \\
     \vspace{2 pt}
     \rotatebox[origin=c]{90}{\tiny{SVD}} & \raisebox{-.5\height}{\includegraphics[width=\mywidth,height=\myheight]{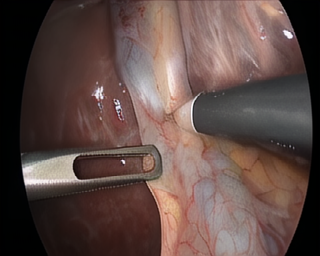}} & \raisebox{-.5\height}{\includegraphics[width=\mywidth,height=\myheight]{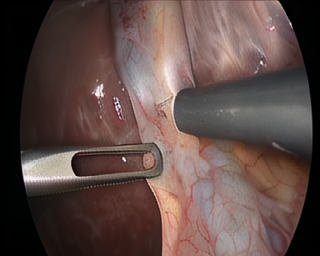}} & \raisebox{-.5\height}{\includegraphics[width=\mywidth,height=\myheight]{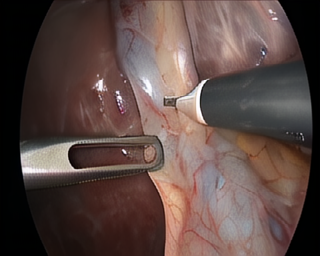}} & \raisebox{-.5\height}{\includegraphics[width=\mywidth,height=\myheight]{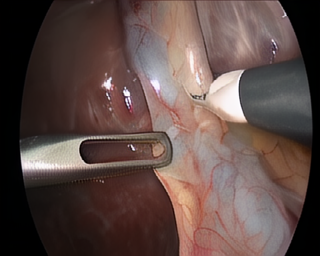}}  & \raisebox{-.5\height}{\includegraphics[width=\mywidth,height=\myheight]{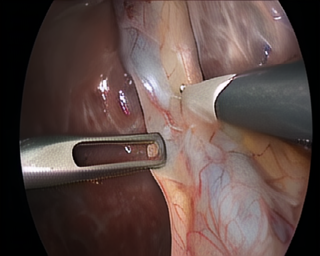}}  & \raisebox{-.5\height}{\includegraphics[width=\mywidth,height=\myheight]{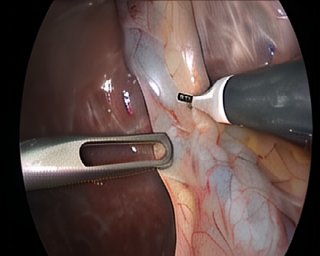}} \\
     
     \rotatebox[origin=c]{90}{\tiny{VISAGE}} & \raisebox{-.5\height}{\includegraphics[width=\mywidth,height=\myheight]{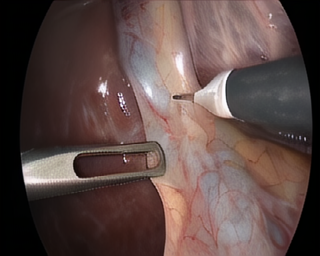}} & \raisebox{-.5\height}{\includegraphics[width=\mywidth,height=\myheight]{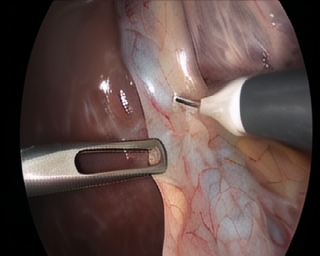}} & \raisebox{-.5\height}{\includegraphics[width=\mywidth,height=\myheight]{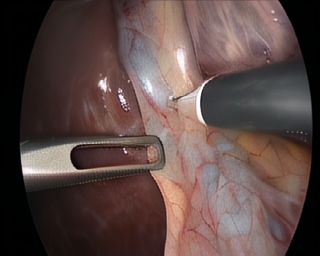}} & \raisebox{-.5\height}{\includegraphics[width=\mywidth,height=\myheight]{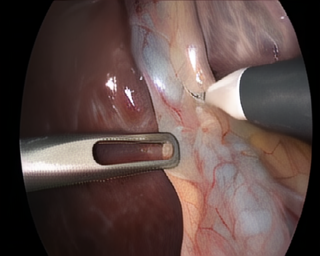}}  & \raisebox{-.5\height}{\includegraphics[width=\mywidth,height=\myheight]{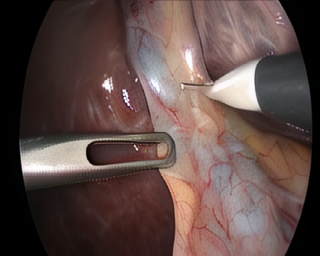}}& \raisebox{-.5\height}{\includegraphics[width=\mywidth,height=\myheight]{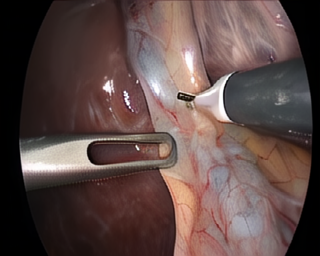}}  \\
\end{tabular}
\caption{\textbf{Qualitative Results.} Comparison of the SOTA models on video synthesis on CholecT50 \cite{nwoye2021rendezvous}.}
    \label{fig:qualitative}
\end{figure}
\vspace{4pt}\noindent\textbf{Qualitative Results}
\autoref{fig:qualitative} demonstrates the qualitative comparison of the SOTA models compared to \methodName{}. CoDi fails to generate realistic results, and Waldo suffers from artifacts in the generated samples. LFDM, SVD, and \methodName{} were able to generate results resembling the real data. However, \methodName{} generates the fine details, such as the tool tip, which is a key component of the surgery, with higher accuracy.

\vspace{4pt}\noindent\textbf{Ablation Study}
We present an ablation study of the different components of \methodName{} in \autoref{tab:ablation}. First, we compare conditioning the model on triplets using a learnable embedding layer versus using the CLIP text embeddings for the action triplet. The results show that utilizing CLIP embedding and further fine-tuning them with the image generation objective achieves the best overall performance. Next, we ablate the fusion of the graph and image embeddings using a linear layer and a cross-attention layer. It can be seen that the introduction of the cross-attention layer largely improves the video generation performance in both conditioning settings.

\section{Conclusion}
In this paper, we have presented a novel task of future video generation in laparoscopic surgery, which can benefit various aspects of surgical data science, such as simulation, analysis, and robot-aided surgery. We have proposed \methodName{}, a generative model that leverages action scene graphs and diffusion models to synthesize realistic and diverse videos of laparoscopic procedures conditioned on a single frame and an action triplet. We have evaluated our method on the CholecT50 dataset, and we have shown that it can generate high-quality and temporally coherent videos that adhere to the domain-specific constraints of laparoscopic surgery. We have also compared our method with several baselines and ablations, and we have demonstrated its superiority in terms of various metrics and qualitative evaluations. Our work opens up new possibilities for future research on surgical video generation, such as incorporating more complex action sequences, improving the diversity and controllability of the generated videos, and exploring the applications of our method in surgical education and training.

%
%
%
\bibliographystyle{splncs04}
\bibliography{ref}

\end{document}